\documentclass[11pt,a4paper]{article}
\usepackage[hyperref]{acl2020}
\usepackage{times}
\usepackage{latexsym}
\usepackage{enumitem}
\usepackage{amssymb}
\usepackage{framed}
\usepackage{amsmath}
\usepackage{bm}

\usepackage{microtype}
\usepackage{amsmath}
\usepackage{booktabs}
\usepackage{multirow, makecell}
\usepackage{longtable}
\usepackage{amsmath}
\usepackage{amsfonts}
\usepackage{graphicx}
\usepackage{algorithm}
\usepackage{algpseudocode}
\usepackage{subcaption}
\usepackage{soul}
\usepackage{caption}
\usepackage{verbatim}
\usepackage{txfonts}

\aclfinalcopy 
\def\aclpaperid{1814} 

\newcommand\drop{\textsc{DROP}}
\newcommand\nlvr{\textsc{NLVR2}}

\newcommand{\LXMERT}{\textsc{LXMERT}}

\newif\ifcomments
\commentstrue
\ifcomments
    \providecommand{\sanjay}[1]{{\protect\color{blue}{[Sanjay: #1]}}}
    \providecommand{\tw}[1]{{\protect\color{magenta}{[TW: #1]}}}
    \providecommand{\jb}[1]{{\protect\color{purple}{[JB: #1]}}}
    \providecommand{\matt}[1]{{\protect\color{teal}{[Matt: #1]}}}
    \providecommand{\nitish}[1]{{\protect\color{violet}{[NG: #1]}}}
    \providecommand{\bb}[1]{{\protect\color{olive}{[BB: #1]}}}
\else
    \providecommand{\sanjay}[1]{}
    \providecommand{\matt}[1]{}
    \providecommand{\jb}[1]{}
    \providecommand{\nitish}[1]{}
    \providecommand{\bb}[1]{}
    \providecommand{\tw}[1]{}
\fi

\title{Obtaining Faithful Interpretations from Compositional Neural Networks}

\author{\makecell{Sanjay Subramanian\thanks{\hspace{0.3em} Equal Contribution}$^{*1}$ ~~~~~~~ Ben Bogin$^{*2}$ ~~~~~~~ Nitish Gupta$^{*3}$ \\ Tomer Wolfson$^{1,2}$ ~~~~~ Sameer Singh$^{4}$ ~~~~~ Jonathan Berant$^{1,2}$ ~~~~~ Matt Gardner$^{1}$ } \\ 
$^{1}$Allen Institute for AI\hspace{5mm}
$^{2}$Tel-Aviv University \\
$^{3}$University of Pennsylvania\hspace{5mm}
$^{4}$University of California, Irvine \\ 
\texttt{\makecell{\{sanjays,mattg\}@allenai.org, \{ben.bogin,joberant\}@cs.tau.ac.il,\\
nitishg@seas.upenn.edu, tomerwol@mail.tau.ac.il, sameer@uci.edu}}}

\date{}

\begin{document}

\setlength{\abovedisplayskip}{3pt}
\setlength{\belowdisplayskip}{3pt}

\maketitle
\begin{abstract}
Neural module networks (NMNs) are a popular approach for modeling compositionality: they achieve high accuracy when applied to problems in language and vision, while  reflecting the compositional structure of the problem in the network architecture.
However, prior work implicitly assumed that the structure of the network modules, describing the abstract reasoning process, provides a faithful explanation of the model's reasoning; that is, that all modules perform their intended behaviour.
In this work, we propose and conduct a systematic evaluation of the intermediate outputs of NMNs on \nlvr{} and \drop{}, two datasets which require composing multiple reasoning steps. We find that the intermediate outputs differ from the expected output, illustrating that the network structure does not provide a faithful explanation of model behaviour.
To remedy that, we train the model with auxiliary supervision and propose particular choices for module architecture that yield much better faithfulness, at a minimal cost to accuracy.
\end{abstract}

\section{Introduction}
\begin{figure}[t]
    \centering
    \includegraphics[width=62mm]{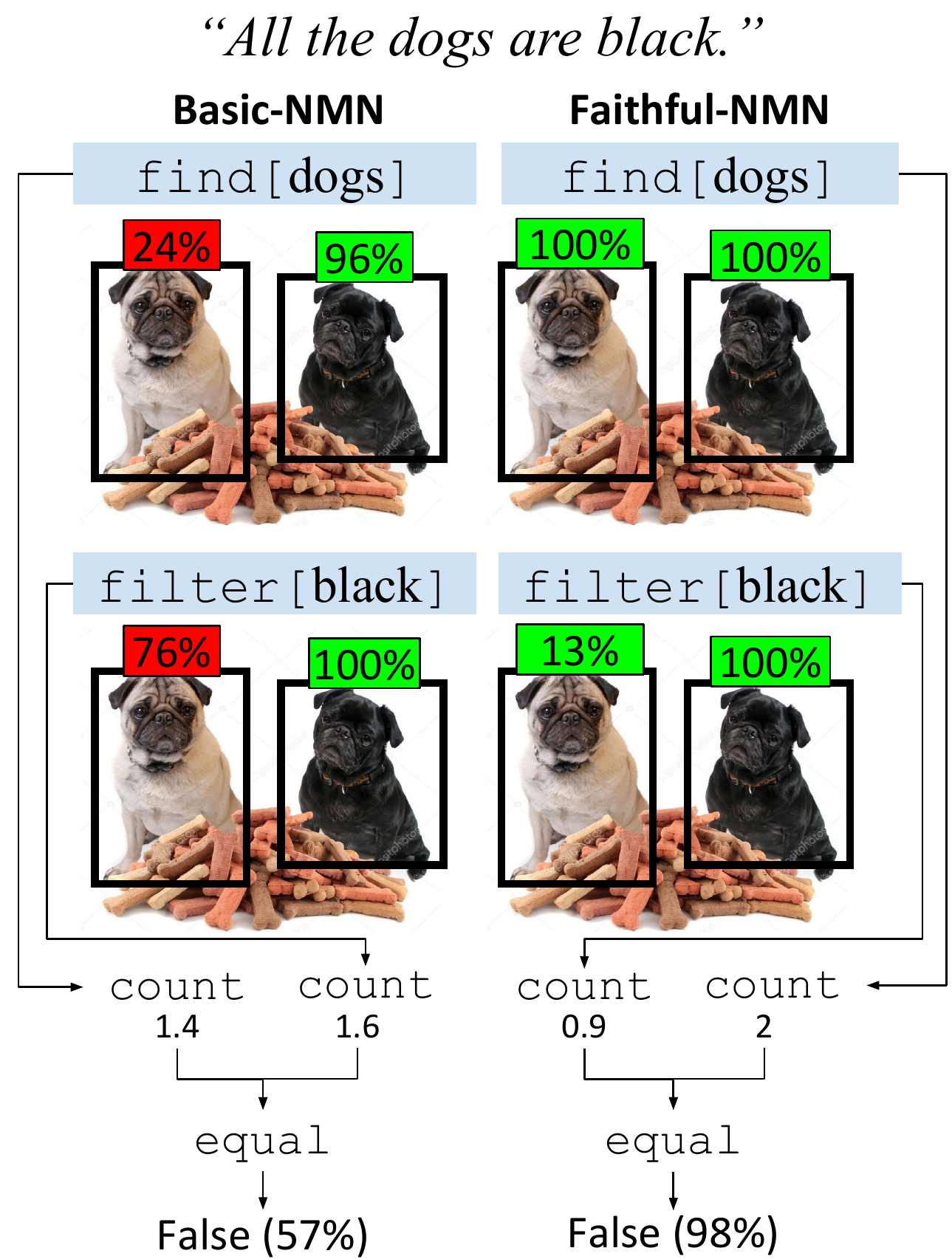}
    \caption{An example for a visual reasoning problem where both the Basic and Faithful NMNs produce the correct answer. 
    The Basic NMN, however, fails to give meaningful intermediate outputs for the \texttt{find} and \texttt{filter} modules, whereas our improved Faithful-NMN assigns correct probabilities in all cases. Boxes are green if probabilities are as expected, red otherwise. }
    \label{fig:intro}
\end{figure}

Models that can read text and reason about it in a particular context (such as an image, a paragraph, or a table) have been recently gaining increased attention, leading to the creation of multiple  datasets that require reasoning in both the visual and textual domain \cite{Johnson2016CLEVRAD, suhr-etal-2017-corpus, talmor-berant-2018-web, yang2018hotpotqa, suhr-etal-2019-corpus, hudson2019gqa, drop}. Consider the example in Figure~\ref{fig:intro} from \nlvr{}: a model must understand the compositional sentence in order to then ground \emph{dogs} in the input, count those that are \emph{black} and verify that the count of all dogs in the image is equal to the number of black dogs.

Both models that assume an intermediate structure \cite{andreas-etal-2016-learning, jiang-bansal-2019-self} and models without such structure \cite{lxmert, hu-etal-2019-multi, min2019compositional} have been proposed for these reasoning problems.
While good performance can be obtained without a structured representation, an advantage of structured approaches is that the reasoning process in such approaches is more \emph{interpretable}. For example, a structured model can explicitly denote that there are two \emph{dogs} in the image, but that one of them is \emph{not black}. Such interpretability improves our scientific understanding, aids in model development, and improves overall trust in a model.

\begin{figure*}[tb]
\centering
    \includegraphics[width=0.87\textwidth]{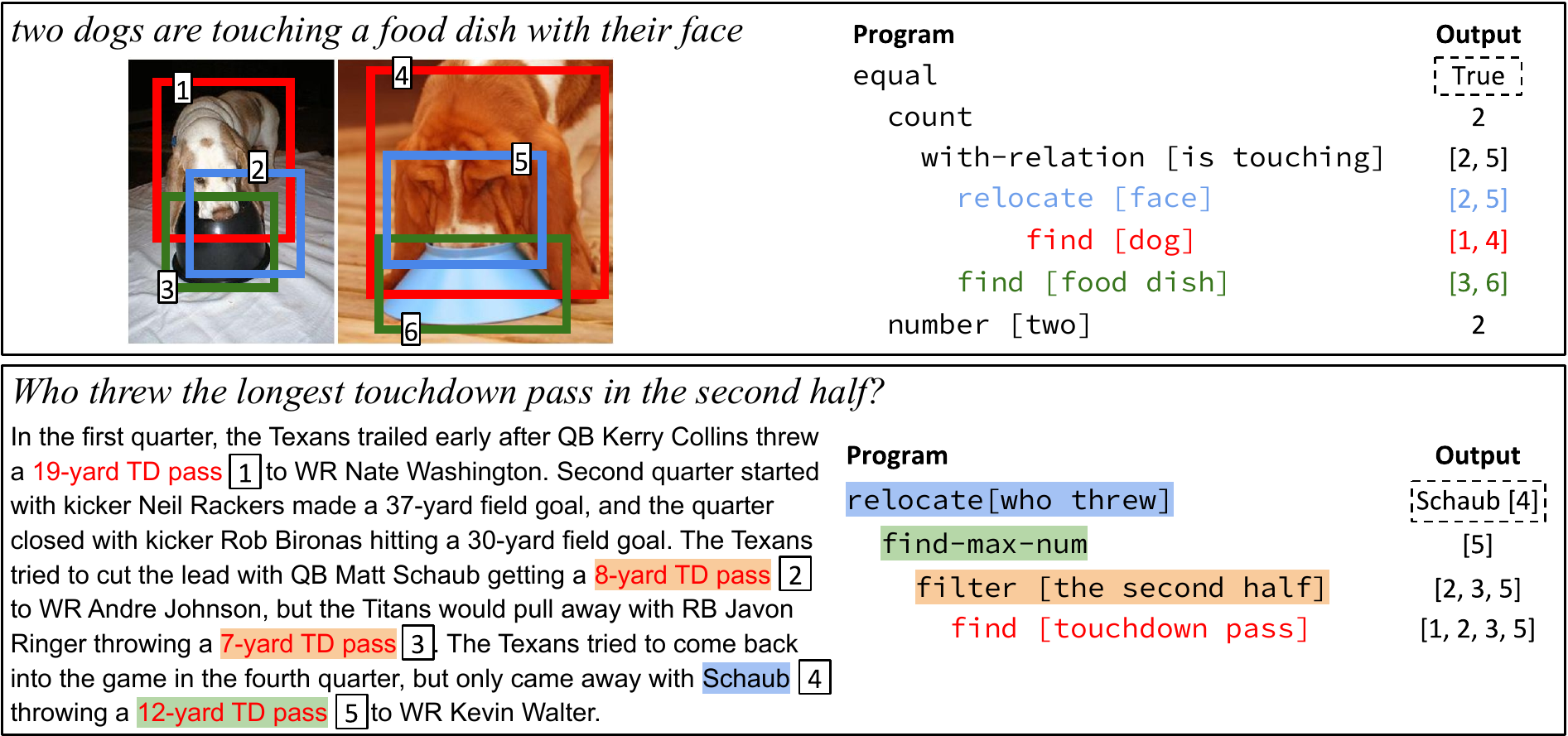}
    \caption{An example for a mapping of an utterance to a gold program and a perfect execution in a reasoning problem from \nlvr{} (top) and \drop{} (bottom).}
 \label{fig:nmn-example}
\end{figure*}

Neural module networks \citep[NMNs;][]{andreas-etal-2016-learning} parse an input utterance into an executable program composed of learnable modules that are designed to perform atomic reasoning tasks and can be composed to perform complex reasoning against an unstructured context. NMNs are appealing since their output is interpretable; they provide a logical meaning representation of the utterance and also the outputs of the intermediate steps (modules) to reach the final answer. 
However, because module parameters are typically learned from end-task supervision only, it is possible that the program will not be a \emph{faithful} explanation of the behaviour of the model \cite{Ross2017RightFT,wiegreffe-pinter-2019-attention}, i.e., the model will solve the task by executing modules according to the program structure, but the modules will not perform the reasoning steps \emph{as intended}.
For example, in Figure~\ref{fig:intro}, a \emph{basic NMN} predicts the correct answer \texttt{False}, but incorrectly predicts the output of the \texttt{find}[\emph{dogs}] operation. It does not correctly locate one of the \emph{dogs} in the image because two of the reasoning steps (\texttt{find} and \texttt{filter}) are collapsed into one module (\texttt{find}). 
This behavior of the \texttt{find} module is not faithful to its intended reasoning operation; a human reading the program would expect \texttt{find}[\emph{dogs}] to \emph{locate} all dogs.
Such unfaithful module behaviour yields an unfaithful explanation of the model behaviour.

Unfaithful behaviour of modules, such as multiple reasoning steps collapsing into one, are undesirable in terms of interpretability; when a model fails to answer some question correctly, it is hard to tell which modules are the sources of error.
While recent work \cite{hu2018explainable, jiang-bansal-2019-self} has shown that one can obtain good performance when using NMNs, the accuracy of individual module outputs was mostly evaluated through qualitative analysis, rather than systematically evaluating the intermediate outputs of each module.

We provide three primary contributions regarding faithfulness in NMNs. 
First, we propose the concept of module-wise faithfulness -- a systematic evaluation of individual module performance in NMNs that judges whether they have learned their intended operations, and define metrics to quantify this for both visual and textual reasoning (\S\ref{sec:measure}).
Empirically, we show on both \nlvr{} \cite{suhr-etal-2019-corpus} and \drop{} \cite{drop} that training a NMN using end-task supervision, even using \emph{gold programs}, does \emph{not} yield module-wise faithfulness, i.e., the modules do not perform their intended reasoning task.
Second, we provide strategies for improving module-wise faithfulness in NMNs (\S\ref{sec:interpret}).
Specifically, (a) we demonstrate how module architecture affects faithfulness (\S\ref{ssec:choice-modules}), 
(b) propose supervising module outputs with either a proxy task or heuristically generated data (\S\ref{ssec:module-sup}), and (c) 
show that providing modules with uncontexualized token representations improves faithfulness (\S\ref{ssec:masking-words}).
Figure~\ref{fig:intro} shows an example where our approach (\emph{Faithful-NMN}) results in expected module outputs as compared to the \emph{Basic-NMN}.
Last, we collect human-annotated intermediate outputs for 536 examples in \nlvr{} and for 215 examples in \drop{} to measure the module-wise faithfulness of models, and publicly release them for future work. Our code and data are available at \url{https://github.com/allenai/faithful-nmn}.
\section{Neural Module Networks}

\paragraph{Overview} 
Neural module networks \cite[NMNs;][]{andreas-etal-2016-learning} are a class of models that map a natural language utterance into an executable program, composed of learnable modules that can be executed against a given context (images, text, etc.), to produce the utterance's denotation (truth value in \nlvr{}, or a text answer in \drop{}). 
Modules are designed to solve atomic reasoning tasks and can be composed to perform complex reasoning.
For example, in Figure~\ref{fig:intro}, the utterance \textit{``All the dogs are black''} is mapped to the program~\texttt{equal(count(find[\textrm{dogs}]), count(filter[\textrm{black}](find[\textrm{dogs}])))}. The \texttt{find} module is expected to find all \textit{dogs} in the image and the \texttt{filter} module is expected to output only the \textit{black} ones from its input. 
Figure~\ref{fig:nmn-example} shows two other example programs with the expected output of each module in the program.

A NMN has two main components: (1) parser, which maps the utterance into an executable program; and (2) executor, which executes the program against the context to produce the denotation. 
In our setup, programs are always trees where each tree node is a module.
In this work, we focus on the executor, and specifically the faithfulness of module execution. We examine NMNs for both text and images, and describe their modules next. 

\subsection{Modules for visual reasoning}
\label{ssec:visual-modules}
In this task, given two images and a sentence that describes the images, the model should output \texttt{True} iff the sentence correctly describes the images. We base our model, the Visual-NMN, on LXMERT  \cite{lxmert}, which takes as input the sentence $x$ and raw pixels, uses Faster R-CNN \cite{faster-rcnn} to propose a set of bounding boxes, $\mathcal{B}$, that cover the objects in the image, and passes the tokens of $x$ and the bounding boxes through a Transformer \cite{vaswani-attention}, encoding the interaction between both modalities. This produces a contextualized representation $\bm{t} \in \mathbb{R}^{|x| \times h}$ for each one of the tokens, and a representation $\bm{v} \in \mathbb{R}^{|\mathcal{B}| \times h}$ for each one of the bounding boxes, for a given hidden dimension $h$.

We provide a full list of modules and their implementation in Appendix \ref{app:modules}.
Broadly, modules take as input representations of utterance tokens through an \emph{utterance attention} mechanism \cite{hu2017learning}, i.e., whenever the parser outputs a module, it also predicts a distribution over the utterance tokens $(p_1, \dots, p_{|x|})$, and the module takes as input $\sum_{i=1}^{|x|} p_i\bm{t}_i$, where $\bm{t}_i$ is the hidden representation of token $i$. In addition, modules produce as output (and take  as input) vectors  $\bm{p} \in \mathbb[0, 1]^{|\mathcal{B}|}$, indicating for each bounding box the probability that it should be output by the module \cite{mao2018the}. 
For example, in the program \texttt{filter}[\emph{black}](\texttt{find}[\emph{dog}]), the \texttt{find} module takes the word `dog' (using \emph{utterance attention}, which puts all probability mass on the word `dog'), and outputs a probability vector $\bm{p}\in \mathbb[0, 1]^{|\mathcal{B}|}$, where ideally all bounding boxes corresponding to dogs have high probability. Then, the \texttt{filter} module takes $\bm{p}$ as input as well as the word `black', and is meant to output high probabilities for bounding boxes with `black dogs'.

For the Visual-NMN we do not use a parser, but rely on a collected set of  \emph{gold} programs (including gold \emph{utterance attention}), as described in \S\ref{sec:experiments}. We will see that despite this advantageous setup, a basic NMN does not produce interpretable outputs.

\subsection{Modules for textual reasoning}
\label{ssec:text-modules}
Our Text-NMN is used to answer questions in the DROP dataset and uses the modules as designed for DROP in prior work~\cite{gupta-nmn-2020} along with three new modules we define in this work.
The modules introduced in \citet{gupta-nmn-2020} and used as is in our Text-NMN are
\texttt{find},
\texttt{filter},
\texttt{relocate},
\texttt{count},
\texttt{find-num}, \texttt{find-date},
\texttt{find-max-num}, \texttt{find-min-num},
\texttt{num-compare} and \texttt{date-compare}.
All these modules are probabilistic and produce, as output, a distribution over the relevant support. For example, \texttt{find} outputs a  distribution over the passage tokens and \texttt{find-num} outputs a distribution over the numbers in the passage.
We extend their model and introduce additional modules; \texttt{addition} and \texttt{subtraction} to add or subtract passage numbers, and \texttt{extract-answer} which directly predicts an answer span from the representations of passage tokens without any explicit compositional reasoning.
We use BERT-base~\cite{devlin-etal-2019-bert} to encode the input question and passage.

The Text-NMN does not have access to gold programs, and thus we 
implement a parser as an encoder-decoder model with attention similar to \citet{krishnamurthy-etal-2017-neural}, which takes the utterance as input, and outputs a linearized abstract syntax tree of the predicted program. 

\section{Module-wise Faithfulness}
\label{sec:measure}
Neural module networks (NMNs) facilitate interpretability of their predictions via the reasoning steps in the structured program and providing the outputs of those intermediate steps during execution. 
For example, in Figure~\ref{fig:nmn-example}, all reasoning steps taken by both the Visual-NMN and Text-NMN can be discerned from the program and the intermediate module outputs.
However, because module parameters are learned from an end-task, there is no guarantee that the modules will learn to perform their intended reasoning operation.
In such a scenario, when modules do not perform their intended reasoning, the program is no longer a faithful explanation of the model behavior since it is not possible to reliably predict the outputs of the intermediate reasoning steps given the program.
Work on NMNs thus far \cite{hu2018explainable, jiang-bansal-2019-self} has overlooked systematically evaluating faithfulness, performing only qualitative analysis of intermediate outputs.

We introduce the concept of \textit{module-wise faithfulness} aimed at evaluating whether each module has correctly learned its intended operation by judging the correctness of its outputs in a trained NMN.
For example, in Figure~\ref{fig:nmn-example} (top), a model would be judged module-wise faithful if the outputs of all the modules, \texttt{find}, \texttt{relocate}, and \texttt{with\_relation}, are correct -- i.e. similar to the outputs that a human would expect.
We provide gold programs when evaluating faithfulness, to not conflate faithfulness with parser accuracy.

\subsection{Measuring faithfulness in Visual-NMN}
\label{ssec:vision-metric}
Modules in Visual-NMN provide for each bounding box a probability for whether it should be a module output. To evaluate intermediate outputs, we sampled examples from the development set, and annotated gold bounding boxes for each instance of \texttt{find}, \texttt{filter}, \texttt{with-relation} and \texttt{relocate}. The annotator draws the correct bounding-boxes for each module in the gold program, similar to the output in Figure \ref{fig:nmn-example} (top).

A module of a faithful model should assign high probability to bounding-boxes that are aligned with the annotated bounding boxes and low probabilities to other boxes. Since the annotated bounding boxes do not align perfectly with the model's bounding boxes, our evaluation must first induce an alignment. 
We consider two bounding boxes as ``aligned'' if the intersection-over-union (IOU) between them exceeds a pre-defined threshold $T=0.5$. 
Note that it is possible for an annotated bounding box to be aligned with several proposed bounding boxes and vice versa.
Next, we consider an \emph{annotated} bounding box $ B_A $ as ``matched'' w.r.t a module output if $ B_A $ is aligned with a proposed bounding box $ B_P $, and $ B_P $ is assigned by the module a probability $>0.5$. Similarly, we consider a \emph{proposed} bounding box $ B_P $ as ``matched'' if $ B_P $ is assigned by the module a probability $>0.5$ and is aligned with some annotated bounding box $ B_A $.

We compute precision and recall for each module type (e.g. \texttt{find}) in a particular example by considering all instances of the module in that example. We define \emph{precision} as the ratio between the number of matched proposed bounding boxes and the number of proposed bounding boxes assigned a probability of more than 0.5. We define \emph{recall} as the ratio between the number of matched annotated bounding boxes and the total number of annotated bounding boxes.\footnote{The numerators of the precision and the recall are different. Please see Appendix ~\ref{app:vis_metric_numerator} for an explanation.} F$_1$ is the harmonic mean of precision and recall. 
Similarly, we compute an ``overall'' precision, recall, and F$_1$ score for an example by considering all instances of all module types in that example. The final score is an average over all examples. Please see Appendix ~\ref{app:vis_metric_average} for further discussion on this averaging. 

\subsection{Measuring faithfulness in Text-NMN}
\label{ssec:text-metric}
Each module in Text-NMN produces a distribution over passage tokens (\S\ref{ssec:text-modules}) which is a soft distributed representation for the selected spans.
To measure module-wise faithfulness in Text-NMN, we
obtain annotations for the set of spans that should be output by each module in the gold program (as seen in Figure~\ref{fig:nmn-example} (bottom)) 
Ideally, all modules (\texttt{find}, \texttt{filter}, etc.) should predict high probability for tokens that appear in the gold spans and \textit{zero} probability for other tokens. 

To measure a module output's correctness, we use a metric akin to cross-entropy loss to measure the deviation of the predicted module output $p_{\text{att}}$ from the gold spans  $S~=~[s_{1}, \ldots, s_{N}]$. Here each span $s_{i} = (t^{i}_{\text{s}}, t^{i}_{\text{e}})$ is annotated as the start and end tokens.
Faithfulness of a module is measured by:
$ I = - \sum_{i=1}^{N}  \Bigg(\log \sum_{j = t^{i}_{\text{s}}}^{t^{i}_{\text{e}}} p_{\text{att}}^{j} \Bigg). $
Lower cross-entropy corresponds to better faithfulness of a module.

\section{Improving Faithfulness in NMNs}
\label{sec:interpret}

Module-wise faithfulness is affected by various factors: the choice of modules and their implementation (\S~\ref{ssec:choice-modules}), use of auxiliary supervision (\S~\ref{ssec:module-sup}), and the use of contextual utterance embeddings (\S~\ref{ssec:masking-words}). We discuss ways of improving faithfulness of NMNs across these dimensions.

\subsection{Choice of modules}
\label{ssec:choice-modules}

\paragraph{Visual reasoning}
The \texttt{count} module always appears in \nlvr{} as one of the top-level modules (see Figures~\ref{fig:intro} and~\ref{fig:nmn-example}).\footnote{Top-level modules are Boolean quantifiers, such as number comparisons like \texttt{equal} (which require \texttt{count}) or \texttt{exist}. We implement \texttt{exist} using a call to \texttt{count} and \texttt{greater-equal} (see Appendix \ref{app:modules}), so \texttt{count} always occurs in the program.} 
We now discuss how its architecture affects faithfulness.
Consider the program, \texttt{count(filter[\textrm{black}](find[\textrm{dogs}]))}. 
Its gold denotation (correct count value) would provide minimal feedback using which the \emph{descendant} modules in the program tree, such as \texttt{filter} and \texttt{find}, need to learn their intended behavior.
However, if \texttt{count} is implemented as an expressive neural network, it might learn to perform tasks designated for \texttt{find} and \texttt{filter}, hurting faithfulness. Thus, an architecture that allows counting, but also encourages \emph{descendant} modules to learn their intended behaviour through backpropagation, is desirable. We discuss three possible \texttt{count} architectures, which take as input the bounding box probability vector $\bm{p} \in [0,1]^{|\mathcal{B}|}$ and the visual features $\bm{v} \in \mathbb{R}^{|\mathcal{B}| \times h}$.

\noindent
\textbf{\emph{Layer-count module}} is motivated by the count architecture of \citet{hu2017learning}, which uses a linear projection from image attention, followed by a \textrm{softmax}. 
This architecture explicitly uses the visual features, $\bm{v}$, giving it greater expressivity compared to simpler methods. 
First we compute $\bm{p}\cdot \bm{v}$, the weighted sum of the visual representations, based on their probabilities, and then output a scalar count using:
$ \text{FF}_1 ( \text{LayerNorm} (  \text{FF}_2 ( \bm{p}\cdot \bm{v} ) ), $
where $\text{FF}_1$ and $\text{FF}_2$ are feed-forward networks, and the activation function of $\text{FF}_1$ is ReLU in order to output positive numbers only.

As discussed, since this implementation has access to the visual features of the bounding boxes, it can learn to perform certain tasks itself, without providing proper feedback to descendant modules. We show in \S\ref{sec:experiments} this indeed hurts faithfulness.

\noindent
\textbf{\emph{Sum-count module}} on the other extreme,
ignores $\bm{v}$, and simply computes the sum $\sum_{i=1}^{|\mathcal{B}|} \bm{p}_i$. Being parameter-less, this architecture provides direct feedback to descendant modules on how to change their output to produce better probabilities. 
However, such a simple functional-form ignores the fact that bounding boxes are overlapping, which might lead to over-counting objects. In addition, 
we would want \texttt{count} to ignore boxes with low probability. For example, if \texttt{filter} predicts a $5\%$ probability for 20 different bounding boxes, we would not want the output of \texttt{count} to be $1.0$.

\noindent
\textbf{\emph{Graph-count module}} \cite{zhang2018learning} is a middle ground between both approaches - the na\"ive \emph{Sum-Count} and the flexible \emph{Layer-Count}. Like \emph{Sum-Count}, it does not use visual features, but learns to ignore overlapping and low-confidence bounding boxes while introducing only a minimal number of parameters (less than $300$). It does so by treating each bounding box as a node in a graph, and then learning to prune edges and cluster nodes based on the amount of overlap between their bounding boxes (see paper for further details).
Because this is a light-weight implementation that does not access visual features, proper feedback from the module can propagate to its descendants, encouraging them to produce better predictions.

\paragraph{Textual reasoning}
In the context of Text-NMN (on \drop{}), we study the effect of several modules on interpretability.

First, we introduce an \texttt{extract-answer} module. This module bypasses all compositional reasoning and directly predicts an answer from the input contextualized representations. This has potential to improve performance, in cases 
where a question describes reasoning that cannot be captured by pre-defined modules, in which case the program can be the \texttt{extract-answer} module only.
However, introducing \texttt{extract-answer} adversely affects interpretability and learning of other modules, specifically in the absence of gold programs. First, \texttt{extract-answer} does not provide any interpretability. Second, whenever the parser predicts the \texttt{extract-answer} module, the parameters of the more interpretable modules are not trained. Moreover, the parameters of the encoder are trained to perform reasoning \emph{internally} in a non-interpretable manner. We study the interpretability vs. performance trade-off by training Text-NMN with and without \texttt{extract-answer}. 

Second, consider the program \texttt{find-max-num(find[\textrm{touchdown}])} that aims to find the \textit{longest touchdown}. 
\texttt{find-max-num} should sort spans by their value and return the maximal one;
if we remove \texttt{find-max-num}, the program would reduce to \texttt{find[\textrm{touchdown}]}, 
and the \texttt{find} module would have to select the longest touchdown rather than all touchdowns, following the true denotation. More generally, omitting atomic reasoning modules pushes other modules to compensate and perform complex tasks that were not intended for them, hurting faithfulness.
To study this, we train Text-NMN by removing sorting and comparison modules (e.g., \texttt{find-max-num} and \texttt{num-compare}), and evaluate how this affects module-wise interpretability.

\subsection{Supervising module output}
\label{ssec:module-sup}
As explained, given end-task supervision only, modules may not act as intended, since their parameters are only trained for minimizing the end-task loss.
Thus, a straightforward way to improve interpretability is to train modules with additional atomic-task supervision.

\paragraph{Visual reasoning}
For Visual-NMN, we pre-train \texttt{find} and \texttt{filter} modules with explicit intermediate supervision, obtained from the GQA balanced dataset \cite{hudson2019gqa}. Note that this supervision is used only during pre-training -- we do not assume we have full-supervision for the actual task at hand.
GQA questions are annotated by gold programs; we focus on 
``exist" questions that use \texttt{find} and \texttt{filter} modules only, such as \emph{``Are there any red cars?"}.

Given gold annotations from Visual Genome \cite{krishna2017visual}, we can compute a label for each of the bounding boxes proposed by Faster-RCNN. We label a proposed bounding box as `positive' if its IOU with a gold bounding box is $>0.75$, and `negative' if it is $<0.25$. We then train on GQA examples, minimizing both the usual denotation loss, as well as an auxiliary loss for each instance of \texttt{find} and \texttt{filter}, which is binary cross entropy for the labeled boxes. This loss rewards high probabilities for `positive' bounding boxes and low probabilities for `negative' ones.

\paragraph{Textual reasoning}
Prior work~\cite{gupta-nmn-2020} proposed heuristic methods to extract  supervision for the \texttt{find-num} and \texttt{find-date} modules in \drop{}. On top of the end-to-end objective, they use an auxiliary objective that encourages these modules to output the ``gold" numbers and dates according to the heuristic supervision. They show that supervising intermediate module outputs helps improve model performance. 
In this work, we evaluate the effect of such supervision on the faithfulness of both the supervised modules, as well as other modules that are trained jointly.

\subsection{Decontextualized word representations}
\label{ssec:masking-words}
The goal of decomposing reasoning into multiples steps, each focusing on different parts of the utterance, is at odds with the widespread use of contextualized representations such as BERT or LXMERT. While the \emph{utterance attention} is meant to capture information only from tokens relevant for the module's reasoning, contextualized token representations carry global information.
For example, consider the program \texttt{filter[\text{\textrm{red}}](find[\textrm{\emph{car}}])} for the phrase \emph{red car}. Even if \texttt{find} attends only to the token \emph{car}, its representation might also express the attribute \emph{red}, so \texttt{find} might learn to find just \emph{red cars}, rather than all \emph{cars}, rendering the \texttt{filter} module useless, and harming faithfulness. To avoid such contextualization in Visual-NMN, we zero out the representations of tokens that are unattended, thus the input to the module is computed (with LXMERT) from the remaining tokens only.

\section{Experiments}

\begin{table*}[t]
\small
\centering
\captionsetup{font=footnotesize}
\resizebox{1.0\textwidth}{!}{
\begin{tabular}{lcccccccc}
\toprule
\multirow{2}[3]{*}{{\bf Model}} & \multirow{2}[3]{*}{\begin{tabular}{@{}c@{}}\textbf{Performance} \\ (Accuracy) \end{tabular}} & \multicolumn{3}{c}{\textbf{Overall Faithful.} ($\uparrow$)} & \multicolumn{4}{c}{Module-wise Faithfulness F$_1$($\uparrow$)} \\
\cmidrule(lr){3-9}
    & & Prec. & Rec. & F$_1$ & find & filter & with-relation & relocate \\
\midrule
LXMERT                               & \textbf{71.7} &      &      &      &      &      \\
\addlinespace
Upper Bound                   &      & 1 & 0.84 &  0.89    &  0.89    &  0.92    &    0.95  &   0.75   \\
\addlinespace
NMN w/ Layer-count                        & 71.2 & 0.39 & 0.39 & 0.11 & 0.12 & 0.20 & 0.37 & \textbf{0.27} \\
NMN w/ Sum-count                          & 68.4 & \textbf{0.49} & 0.31 & 0.28 & 0.31 & 0.32 & 0.44 & 0.26 \\
NMN w/ Graph-count                        & 69.6 & 0.37 & 0.39 & 0.28 & 0.31 & 0.29 & 0.37 & 0.19 \\
\addlinespace

NMN w/  Graph-count + decont.                   & 67.3 & 0.29 & 0.51 & 0.33 & 0.38 & 0.30 & 0.36 & 0.13 \\

\addlinespace

NMN w/ Graph-count + pretraining            & 69.6 & 0.44 & 0.49 & 0.36 & 0.39 & 0.34 & 0.42 & 0.21 \\
\addlinespace

NMN w/ Graph-count + decont. + pretraining        & 68.7 & 0.42 & \textbf{0.66} & \textbf{0.47} & \textbf{0.52} & \textbf{0.41} & \textbf{0.47} & 0.21 \\
\bottomrule
\end{tabular}
}
\caption{Faithfulness and accuracy on \nlvr{}. ``decont.'' refers to decontextualized word representations. Precision, recall, and F$_1$ are averages across examples, and thus F$_1$ is \textbf{not} the harmonic mean of the corresponding precision and recall.}
\label{tab:nlvr-results}
\end{table*}

\begin{table*}[tbh]
\small
\centering
\captionsetup{font=footnotesize}
\resizebox{1.0\textwidth}{!}{
\begin{tabular}{lccccccc}
\toprule
\multirow{2}[3]{*}{\textbf{Model}} & 
\multirow{2}[3]{*}{\begin{tabular}{@{}c@{}}\textbf{Performance} \\ (F$_1$ Score)\end{tabular}} & \multirow{2}[3]{*}{\begin{tabular}{@{}c@{}}\textbf{Overall Faithful.} \\ (cross-entropy$^{*}$ $\downarrow$) \end{tabular}} & 
\multicolumn{5}{c}{Module-wise Faithfulness$^{*}$ ($\downarrow$)} \\
\cmidrule(lr){4-8}
                            &    &                & find  & filter & relocate & min-max$^\dagger$ & find-arg$^\dagger$ \\
\midrule
Text-NMN w/o prog-sup     &      &      &       &      &      &      &       \\
\;\;\; w/  \texttt{extract-answer}      & 63.5  & 9.5          & 13.3  & 9.5  & 3.5  & 2.6  & 9.9 \\
\;\;\; w/o \texttt{extract-answer}      & 60.8  & \textbf{6.9} & 8.1   & 7.3  & 1.3  & 1.7  & 8.5  \\

\addlinespace

Text-NMN w/ prog-sup     &      &      &       &      &      &      &       \\
\;\; no auxiliary sup           & 65.3          &  11.2         &  13.7         &  16.9          &  1.5          &  2.2          &  13.0         \\
\;\; w/o sorting \& comparison  & 63.8          &  8.4          &  9.6          &  11.1          &  1.6          &  1.3          &  10.6         \\
\;\; w/ module-output-sup       & \textbf{65.7} &  \textbf{6.5} &  \textbf{7.6} &  \textbf{10.7} &  \textbf{1.3} &  \textbf{1.2} &  \textbf{7.6} \\
\bottomrule
\end{tabular}
}
    \caption{Faithfulness and performance scores for various NMNs on \drop{}.
    $^*$lower is better. $^\dagger$min-max is average faithfulness of \texttt{find-min-num} and \texttt{find-max-num}; find-arg of \texttt{find-num} and \texttt{find-date}.}
    \label{tab:drop-results}
\end{table*}

\label{sec:experiments}
We first introduce the datasets used and the experimental setup for measuring faithfulness (\S~\ref{ssec:exp-setup}).
We demonstrate that training NMNs using end-task supervision only does not yield module-wise faithfulness both for visual and textual reasoning. We then show that the methods from \S\ref{sec:interpret} are crucial for achieving faithfulness and how different design choices affect it (\S~\ref{ssec:exp-results}).
Finally, we qualitatively show examples of improved faithfulness and analyze possible reasons for errors  (\S~\ref{ssec:exp-analysis}).

\subsection{Experimental setup}
\label{ssec:exp-setup}
Please see Appendix \ref{app:tr-details} for further detail about the experimental setups.

\paragraph{Visual reasoning}
We automatically generate gold program annotations for $26,311$ training set examples and for $5,772$ development set examples from \nlvr{}. The input to this generation process is the set of crowdsourced question decompositions from the \textsc{Break} dataset \cite{Wolfson2020Break}. See Appendix \ref{app:annotation} for details.
For module-wise faithfulness evaluation, 536 examples from the development set were annotated with the gold output for each module by experts.

\paragraph{Textual reasoning}
We train Text-NMN on \drop{}, which is augmented with program supervision for $4,000$ training questions collected heuristically as described in \citet{gupta-nmn-2020}.
The model is evaluated on the complete development set of DROP which does not contain any program supervision.
Module-wise faithfulness is measured on $215$ manually-labeled questions from the development set, which are annotated with gold programs and module outputs (passage spans).

\subsection{Faithfulness evaluation}
\label{ssec:exp-results}
\paragraph{Visual reasoning}
Results are seen in Table \ref{tab:nlvr-results}. Accuracy for \LXMERT, when trained and evaluated on the same subset of data, is 71.7\%; slightly higher than NMNs, but without providing evidence for the compositional structure of the problem.

For faithfulness, we measure an upper-bound on the faithfulness score. Recall that this score measures the similarity between module outputs and annotated outputs. Since module outputs are constrained by the bounding boxes proposed by Faster-RCNN (\S\ref{ssec:visual-modules}), while annotated boxes are not, perfect faithfulness could only be achieved by a model if there are suitable bounding boxes. \emph{Upper Bound} shows the maximal faithfulness score conditioned on the proposed bounding boxes.

We now compare the performance and faithfulness scores of the different components.
When training our NMN with the most flexible count module, (\emph{NMN w/ Layer-count}), an accuracy of $71.2\%$ is achieved, a slight drop compared to LXMERT but with low faithfulness scores. Using \emph{Sum-count} drops about $3\%$ of performance, but increases faithfulness. Using \emph{Graph-count} increases accuracy while faithfulness remains similar.

Next, we analyze the effect of decontextualized word representations (abbreviated ``decont.'') and pre-training. First, we observe that \emph{NMN w/ Graph-count + decont.} increases faithfulness score to $0.33$ F$_1$ at the expense of accuracy, which drops to $67.3\%$. Pre-training (\emph{NMN w/ Graph-count + pretraining}) achieves higher faithfulness scores with a higher accuracy of $69.6\%$. Combining the two achieves the best faithfulness ($0.47$ F$_1$) with a minimal accuracy drop. We perform a paired permutation test to compare \emph{NMN w/ Graph-count + decont. + pretraining} with \emph{NMN w/ Layer-count} and find that the difference in F$_1$ is statistically significant ($ p < 0.001 $). Please see Appendix ~\ref{app:sig-test-visual} for further details.

\paragraph{Textual reasoning}   
As seen in Table~\ref{tab:drop-results}, when trained on \drop{} using question-program supervision, the model achieves $65.3$ F$_1$ performance and a faithfulness score of $11.2$.
When adding supervision for intermediate modules (\S\ref{ssec:module-sup}), we find that the module-wise faithfulness score improves to $6.5$.
Similar to Visual-NMN, this shows that supervising intermediate modules in a program leads to better faithfulness.

To analyze how choice of modules affects faithfulness, we train without sorting and comparison modules  (\texttt{find-max-num}, \texttt{num-compare}, etc.). We find that while performance drops slightly, faithfulness deteriorates significantly to $8.4$,
showing that modules that perform atomic reasoning are crucial for faithfulness.
When trained without program supervision, removing \texttt{extract-answer} improves faithfulness  ($9.5\rightarrow 6.9$) but at the cost of performance ($63.5 \rightarrow 60.8$ F$_1$).
This shows that such a black-box module encourages reasoning in an opaque manner, but can improve performance by overcoming the limitations of pre-defined modules.
All improvements in faithfulness are significant as measured using paired permutation tests ($ p < 0.001 $).

\paragraph{Generalization} A natural question is whether models that are more faithful also generalize better. We conducted a few experiments to see whether this is true for our models. For NLVR2, we performed (1) an experiment in which programs in training have length at most $7$, and programs at test time have length greater than $7$, (2) an experiment in which programs in training have at most $1$ \texttt{filter} module and programs at test time have at least $2$ \texttt{filter} modules, and (3) an experiment in which programs in training do not have both \texttt{filter} and \texttt{with-relation} modules in the same program, while each program in test has both modules. We compared three of our models -- \emph{NMN w/ Layer-count}, \emph{NMN w/ Sum-count}, and \emph{NMN w/ Graph-count + decont. + pretraining}. We did not observe that faithful models generalize better (in fact, the most unfaithful model tended to achieve the best generalization). 

To measure if faithful model behavior leads to better generalization in Text-NMN we conducted the following experiment. We selected the subset of data for which we have gold programs and split the data such that questions that require maximum and greater-than operations are present in the training data while questions that require computing minimum and less-than are in the test data.
We train and test our model by providing gold-programs under two conditions, in the presence and absence of additional module supervision. We find that providing auxiliary module supervision (that leads to better module faithfulness; see above) also greatly helps in model generalization (performance increases from $32.3$ F$_1$ $\rightarrow$ $78.3$ F$_1$).

\subsection{Qualitative analysis}
\label{ssec:exp-analysis}

\begin{figure}[t]
    \centering
    \includegraphics[width=75mm]{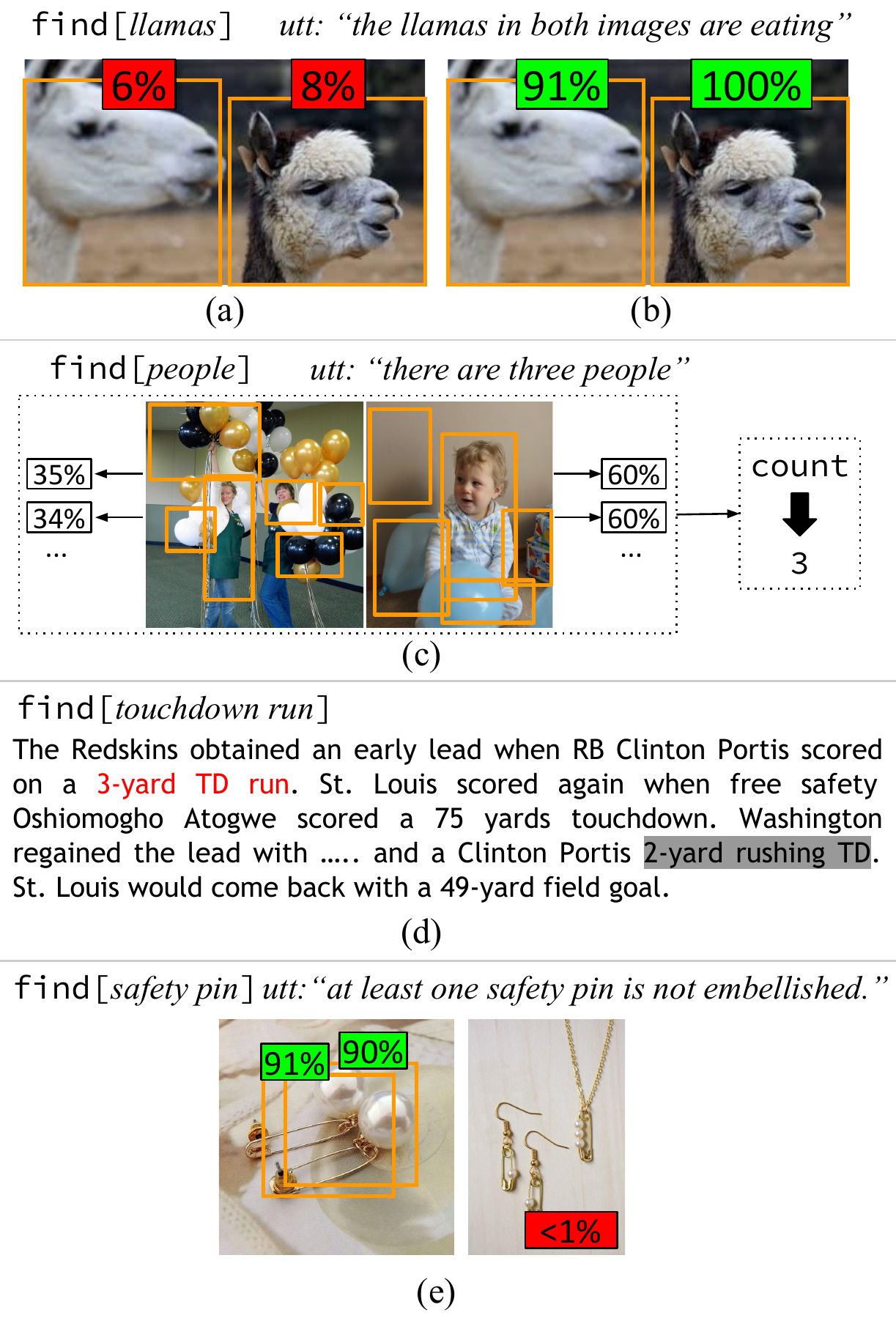}
    \caption{Comparison of module outputs between NMN versions: (a) Visual-NMN with contextualized representations, (b) Visual-NMN with decontextualized representations, (c) model using a parameter-rich count layer (Layer-Count), (d) Text-NMN trained without sorting module produces an incorrect \texttt{find} output (misses \textit{2-yard rushing TD}), and (e) Visual-NMN failure case with a rare object (of \emph{w/ Graph-count + decont. + pretraining})}
    \label{fig:output_compare}
\end{figure}

\paragraph{Model comparisons} 
We analyze outputs of different modules in Figure~\ref{fig:output_compare}.
Figures \ref{fig:output_compare}a, \ref{fig:output_compare}b show the output of \texttt{find}[\emph{llamas}] when trained with contextualized and decontextualized word representations. With contextualized representations (\ref{fig:output_compare}a), the \texttt{find} fails to select any of the \emph{llamas}, presumably because it can observe the word \emph{eating}, thus effectively searching for \emph{eating llamas}, which are not in the image. Conversely, the decontextualized model correctly selects the boxes.
Figure \ref{fig:output_compare}c shows that \texttt{find} outputs meaningless probabilities for most of the bounding boxes when trained with \emph{Layer-count}, yet the count module produces the correct value (three). 
Figure~\ref{fig:output_compare}d shows that \texttt{find} fails to predict all relevant spans when trained without sorting modules in Text-NMN.

\paragraph{Error analysis}
We analyze cases where outputs were unfaithful. First, for visual reasoning, we notice that faithfulness scores are lower for long-tail objects. For example, for \emph{dogs}, a frequent noun in \nlvr{}, the execution of \texttt{find[\textrm{\emph{dogs}]}} yields an average faithfulness score of 0.71, while items such as \emph{roll of toilet paper}, \emph{barbell} and \emph{safety pin} receive lower scores (0.22, 0.29 and 0.05 respectively; example for a failure case for \emph{safety pin} in Fig.~\ref{fig:output_compare}e). In addition, some objects are harder to annotate with a box (\emph{water}, \emph{grass}, \emph{ground}) and therefore receive low scores. The issue of small objects can also explain the low scores of \texttt{relocate}. In the gold box annotations used for evaluation, the average areas for \texttt{find}, \texttt{filter}, \texttt{with-relation}, and \texttt{relocate} (as a fraction of the total image area) are $ 0.19 $, $ 0.19 $, $ 0.15 $, and $ 0.07 $, respectively. Evidently, \texttt{relocate} is executed with small objects that are harder to annotate (\emph{tongue}, \emph{spots}, \emph{top of}), and indeed the upper-bound and model scores for \texttt{relocate} are lowest among the module types.

\section{Related Work}
NMNs were originally introduced for visual question answering and applied to datasets with synthetic language and images as well as VQA \cite{VQA}, whose questions require few reasoning steps \cite{andreas-etal-2016-learning, hu2017learning, hu2018explainable}. In such prior work, module-wise faithfulness was mostly assessed via qualitative analysis of a few examples \cite{jiang-bansal-2019-self, gupta-nmn-2020}. \citet{hu2018explainable} did an evaluation where humans rated the clarity of the reasoning process and also tested whether humans could detect model failures based on module outputs. In contrast, we quantitatively measure each module's predicted output against the annotated gold outputs.

A related systematic evaluation of interpretability in VQA was conducted by \citet{interpretableCounting}. They evaluated the interpretability of their VQA counting model, where the interpretability score is given by the semantic similarity between the gold label for a bounding box and the relevant word(s) in the question. However, they studied only counting questions, which were also far less compositional than those in \nlvr{} and \drop{}.

Similar to the gold module output annotations that we provide and evaluate against, \textsc{HotpotQA} \cite{yang2018hotpotqa} and \textsc{CoQA} \cite{reddy2019coqa} datasets include supporting facts or rationales for the answers to their questions, which can be used for both supervision and evaluation.

In concurrent work, \citet{jacovi2020towards} recommend studying faithfulness on a scale rather than as a binary concept. Our evaluation method can be viewed as one example of this approach.

\section{Conclusion}
We introduce the concept of \emph{module-wise faithfulness}, a systematic evaluation of faithfulness in neural module networks (NMNs) for visual and textual reasoning. We show that na\"ive training of NMNs does not produce faithful modules and propose several techniques to improve module-wise faithfulness in NMNs.
We show how our approach leads to much higher module-wise faithfulness at a low cost to performance. We encourage future work to judge model interpretability using the proposed evaluation and publicly published annotations, and explore techniques for improving faithfulness and interpretability in compositional models.

\section*{Acknowledgements}
We thank members of UCI NLP, TAU NLP, and the AllenNLP teams as well as Daniel Khashabi for comments on earlier drafts of this paper. We also thank the anonymous reviewers for their comments. This research was partially supported
by The Yandex Initiative for Machine Learning, the European Research Council (ERC) under the European Union Horizons 2020 research and innovation programme (grant ERC DELPHI 802800), 
funding by the ONR under Contract No. N00014-19-1-2620, 
and by sponsorship from the LwLL DARPA program under Contract No. FA8750-19-2-0201. 
This work was completed in partial fulfillment for the Ph.D degree of Ben Bogin.
\pagebreak
\bibliography{acl2020}
\bibliographystyle{acl_natbib}

\clearpage

\appendix

\begin{table*}[t]
\centering
\begin{tabular}{llrl}
\hline \textbf{Module} & \textbf{Output} & \textbf{Implementation} \\ \hline 
\texttt{find}$ [q_{att}] $ & $p$ & $ W_1^T([x;v])+b_1 $ \\
\texttt{filter}$[q_{att}](p)$ & $p$ &$ p \odot (W_1^T([x;v])+b_1) $ \\
\texttt{with-relation}$[q_{att}](p_1, p_2)$ & $p$ &$\max(p_2)p_1 \odot \text{MLP}([x;v_1;v_2])$ \\
\texttt{project}$[q_{att}](p)$ & $p$ &$\max(p)\texttt{find}(q_{att}) \odot \text{MLP}([W_2;v_1;v_2])$ \\
\texttt{count}$(p)$ & N & \texttt{number}$ \left(\sum(p), \sigma^2\right) $ \\
\hline
\texttt{exist}$(p)$ & B & \texttt{greater-equal}(p, 1) \\
\texttt{greater-equal} $(a: \textrm{N}, b: \textrm{N})$ & B & \texttt{greater}(a, b) + \texttt{equal}(a, b) \\
\texttt{less-equal} $(a: \textrm{N}, b: \textrm{N})$ & B &  \texttt{less}(a, b) + \texttt{equal}(a, b) \\
\texttt{equal}$(a: \textrm{N}, b: \textrm{N})$ & B &  $ \sum_{k=0}^{K} \Pr[a = k]\Pr[b = k]$\\
\texttt{less}$(a: \textrm{N}, b: \textrm{N})$ & B &  $ \sum_{k=0}^{K} \Pr[a = k]\Pr[b > k]$\\
\texttt{greater}$(a: \textrm{N}, b: \textrm{N})$ & B &  $ \sum_{k=0}^{K} \Pr[a = k]\Pr[b < k] $\\
\texttt{and}$(a: \textrm{B}, b: \textrm{B})$ & B & a*b \\
\texttt{or}$(a: \textrm{B}, b: \textrm{B})$ & B & a+b-a*b \\
\texttt{number}$(m: \textrm{F}, v: \textrm{F})$ & N & Normal$ (\textrm{mean}=m, \textrm{var}=v) $\\
\texttt{sum}$(a: \textrm{N}, b: \textrm{N})$ & N & \texttt{number} $(a_{mean}+b_{mean}, a_{var}+b_{var}) $\\
\texttt{difference}$(a: \textrm{N}, b: \textrm{N})$ & N &  \texttt{number} $(a_{mean}-b_{mean}, a_{var}+b_{var}) $\\
\texttt{division}$(a: \textrm{N}, b: \textrm{N})$ & N &  \texttt{number} $\left(\frac{a_{mean}}{b_{mean}}+\frac{b_{var}a_{mean}}{b_{mean}^3}, \frac{a_{mean}^2}{b_{mean}^2}\left(\frac{a_{var}}{a_{mean}^2}+\frac{b_{var}}{b_{mean}^2}\right)\right) $\\
\texttt{intersect}$(p_1, p_2)$ & $p$ & $p_1 \cdot p_2 $ \\
\texttt{discard}$(p_1, p_2)$ & $p$ & $ \max(p_1-p_2, 0) $\\
\texttt{in-left-image}$(p)$ & $p$ & $p$ s.t. probabilities for right image are $0$ \\
\texttt{in-right-image}$(p)$ & $p$ & $p$ s.t. probabilities for left image are $0$ \\
\texttt{in-at-least-one-image} & B & macro (see caption) \\
\texttt{in-each-image} & B & macro (see caption) \\
\texttt{in-one-other-image} & B & macro (see caption) \\
\hline
\end{tabular}
\caption{\label{nlvr2-modules-table} Implementations of modules for NLVR2 NMN. First five contain parameters, the rest are deterministic. The implementation of \texttt{count} shown here is the Sum-count version; please see Section \ref{sec:interpret} for a description of other count module varieties and a discussion of their differences. `B' denotes the Boolean type, which is a probability value ([0..1]). `N' denotes the Number type which is a probability distribution. $ K = 72 $ is the maximum count value supported by our model. To obtain probabilities, we first convert each Normal random variable $ X $ to a categorical distribution over $ \{0, 1, ..., K\} $ by setting $ \Pr[X = k] = \Phi(k+0.5)-\Phi(k-0.5) $ if $ k \in \{1, 2, ..., K-1\} $. We set $ \Pr[X = 0] = \Phi(0.5) $ and $ \Pr[X = K] = 1-\Phi(K-0.5) $. Here $ \Phi(\cdot) $ denotes the cumulative distribution function of the Normal distribution. $ W_1 $, $ W_2 $ are weight vectors with shapes $ 2h\times 1 $ and $ h\times 1$, respectively. Here $ h = 768 $ is the size of LXMERT's representations. $ b_1 $ is a scalar weight. $ \text{MLP} $ denotes a two-layer neural network with a GeLU activation \cite{gelu} between layers. $ x $ denotes a question representation, and $ v_i $ denotes encodings of objects in the image. $ x $ and $ v_i $ have shape $ h \times |\mathcal{B}| $, where $ |\mathcal{B}| $ is the number of proposals. $ p $ denotes a vector of probabilities for each proposal and has shape $ 1 \times |\mathcal{B}| $. $ \odot $ and $ [;] $ represent elementwise multiplication and matrix concatenation, respectively. The expressions for the mean and variance in the division module are based on the approximations in \citet{divisionFormula}. The macros execute a given program on the two input images. \texttt{in-at-least-one-image} macro returns true iff the program returns true when executed on at least one of the images. \texttt{in-each-image} returns true iff the program returns true when executed on both of the images. \texttt{in-one-other-image} takes two programs and returns true iff one program return true on left image and second program returns true on right image, or vice-versa.}
\label{tab:modules}
\end{table*}

\section{Modules}
\label{app:modules}
We list all modules for Visual-NMN in Table \ref{tab:modules}.

For Text-NMN, as mentioned, we use all modules are described in \citet{gupta-nmn-2020}. In this work, we introduce the (a) \texttt{addition} and \texttt{subtraction} modules that take as input two distributions over numbers mentioned in the passage and produce a distribution over all posssible addition and subtraction values possible. The output distribution here is the expected distribution for the random variable $Z = X + Y$ (for \texttt{addition}), and (b) \texttt{extract-answer} that produces two distributions over the passage tokens denoting the probabilities for the start and end of the answer span. This distribution is computed by mapping the passage token representations using a simple MLP and softmax operation.

\section{Measuring Faithfulness in Visual-NMN}
\subsection{Numerators of Precision and Recall}
\label{app:vis_metric_numerator}
As stated in Section ~\ref{ssec:vision-metric}, for a given module type and a given example, precision is defined as the number of matched proposed bounding boxes divided by the number of proposed bounding boxes to which the module assigns a probability more than 0.5. Recall is defined as the number of matched annotated bounding boxes divided by the number of annotated bounding boxes. Therefore, the numerators of the precision and the recall need not be equal. In short, the reason for the discrepancy is that there is no one-to-one alignment between annotated and proposed bounding boxes. To further illustrate why we chose not to have a common numerator, we will consider two sensible choices for this shared numerator and explain the issues with them.

One choice for the common numerator is the number of matched proposed bounding boxes. If we were to keep the denominator of the recall the same, then the recall would be defined as the number of matched proposed bounding boxes divided by the number of annotated bounding boxes. Consider an example in which there is a single annotated bounding box that is aligned with five proposed bounding boxes. When this definition of recall is applied to this example, the numerator would exceed the denominator. Another choice would be to set the denominator to be the number of proposed bounding boxes that are aligned with some annotated bounding box. In the example, this approach would penalize a module that gives high probability to only one of the five aligned proposed bounding boxes. However, it is not clear that a module giving high probability to all five proposed boxes is more faithful than a module giving high probability to only one bounding box (e.g. perhaps one proposed box has a much higher IOU with the annotated box than the other proposed boxes). Hence, this choice for the numerator does not make sense.

Another choice for the common numerator is the number of matched annotated bounding boxes. If we were to keep the denominator of the precision the same, then the precision would be defined as the number of matched annotated bounding boxes divided by the number of proposed bounding boxes to which the module assigns probability more than 0.5. Note that since a single proposed bounding box can align with multiple annotated bounding boxes, it is possible for the numerator to exceed the denominator.

Thus, these two choices for a common numerator have issues, and we avoid these issues by defining the numerators of precision and recall separately.
\subsection{Averaging Faithfulness Scores}
\label{app:vis_metric_average}
The method described in Section ~\ref{ssec:vision-metric} computes a precision, recall, and F$_1$ score for each example for every module type occurring in that example. The faithfulness scores reported in Table ~\ref{tab:nlvr-results} are averages across examples. We also considered two other ways of aggregating scores across examples:
\begin{enumerate}
    \item Cumulative P/R/F1: For each module type, we compute a single cumulative precision and recall across all examples. We then compute the dataset-wide F$_1$ score as the harmonic mean of the precision and the recall. The results using this method are in Table ~\ref{tab:nlvr-results-method-1}. There are some differences between these results and those in Table ~\ref{tab:nlvr-results}, e.g. in these results, \emph{NMN w/ Graph-count + decont. + pretraining} has the highest faithfulness score for every module type, including \texttt{relocate}.
    \item Average over module occurrences: For each module type, for each occurrence of the module we compute a precision and recall and compute F$_1$ as the harmonic mean of precision and recall. Then for each module type, we compute the overall precision as the average precision across module occurrences and similarly compute the overall recall and F$_1$. Note that a module can occur multiple times in a single program and that each image is considered a separate occurrence. The results using this method are in Table ~\ref{tab:nlvr-results-method-2}. Again, there are some differences between these results and those in Table ~\ref{tab:nlvr-results}, e.g. \emph{NMN w/ Sum-count} has a slightly higher score for \texttt{with-relation} than \emph{NMN w/ Graph-count + decont. + pretraining}.
\end{enumerate}
With both of these alternative score aggregation methods, we still obtained $ p < 0.001 $ in our significance tests.

We also noticed qualitatively that the metric can penalize modules that assign high probability to proposed bounding boxes that have a relatively high IOU that does not quite pass the IOU threshold of $ 0.5 $. In such cases, while it may not make sense to give the model credit in its recall score, it also may not make sense to penalize the model in its precision score. Consequently, we also performed an evaluation in which for the precision calculation we set a separate ``negative'' IOU threshold of $ 10^{-8} $ (effectively $ 0 $) and only penalized modules for high probabilities assigned to proposed boxes whose IOU is below this threshold. The results computed with example-wise averaging are provided in Table ~\ref{tab:nlvr-results-iou-minus-1e-8}.
\begin{table*}[t]
\small
\centering
\captionsetup{font=footnotesize}
\resizebox{1.0\textwidth}{!}{
\begin{tabular}{lcccccccc}
\toprule
\multirow{2}[3]{*}{{\bf Model}} & \multirow{2}[3]{*}{\begin{tabular}{@{}c@{}}\textbf{Performance} \\ (Accuracy) \end{tabular}} & \multicolumn{3}{c}{\textbf{Overall Faithful.}($\uparrow$)}    & \multicolumn{4}{c}{Module-wise Faithfulness($\uparrow$)} \\
\cmidrule(lr){3-9}
    &  & Prec. & Rec. & F1 & find & filter & with-relation & relocate \\
\midrule
LXMERT                               & \textbf{71.7} &      &      &      &      &      \\
\addlinespace
Upper Bound                   &      & 1 & 0.63 & 0.77    &  0.78    &  0.79    &    0.73  &   0.71   \\
\addlinespace
NMN w/ Layer-count                        & 71.2 & 0.069 & 0.29 & 0.11 & 0.13 & 0.09 & 0.07 & 0.05 \\
NMN w/ Sum-count                          & 68.4 & 0.25 & 0.18 & 0.21 & 0.23 & 0.20 & 0.16 & 0.05 \\
NMN w/ Graph-count                        & 69.6 & 0.20 & 0.22 & 0.21 & 0.24 & 0.19 & 0.17 & 0.04 \\
\addlinespace
NMN w/  Graph-count + decont.                   & 67.3 & 0.21 & 0.29 & 0.24 & 0.28 & 0.22 & 0.19 & 0.04 \\

\addlinespace
NMN w/ Graph-count + pretraining            & 69.6 & 0.28 & 0.31 & 0.30 & 0.34 & 0.27 & 0.25 & 0.09 \\
\addlinespace
NMN w/ Graph-count + decont. + pretraining        & 68.7 & \textbf{0.34} & \textbf{0.43} & \textbf{0.38} & \textbf{0.43} & \textbf{0.34} & \textbf{0.29} & \textbf{0.11} \\
\bottomrule
\end{tabular}
}
\caption{Faithfulness scores on \nlvr{} using the cumulative precision/recall/F$_1$ evaluation.}
\label{tab:nlvr-results-method-1}
\end{table*}
\begin{table*}[t]
\small
\centering
\captionsetup{font=footnotesize}
\resizebox{1.0\textwidth}{!}{
\begin{tabular}{lcccccccc}
\toprule
\multirow{2}[3]{*}{{\bf Model}} & \multirow{2}[3]{*}{\begin{tabular}{@{}c@{}}\textbf{Performance} \\ (Accuracy) \end{tabular}} & \multicolumn{3}{c}{\textbf{Overall Faithful.}($\uparrow$)}    & \multicolumn{4}{c}{Module-wise Faithfulness($\uparrow$)} \\
\cmidrule(lr){3-9}
    &  & Prec. & Rec. & F1 & find & filter & with-relation & relocate \\
\midrule
LXMERT                               & \textbf{71.7} &      &      &      &      &      \\
\addlinespace
Upper Bound                   &      & 1 & 0.91 & 0.92    &  0.90    &  0.95    &    0.96  &   0.82   \\
\addlinespace
NMN w/ Layer-count                        & 71.2 & 0.67 & 0.64 & 0.39 & 0.21 & 0.50 & 0.61 & \textbf{0.50} \\
NMN w/ Sum-count                          & 68.4 & \textbf{0.70} & 0.59 & 0.48 & 0.38 & 0.53 & \textbf{0.63} & 0.49 \\
NMN w/ Graph-count                        & 69.6 & 0.55 & 0.64 & 0.43 & 0.36 & 0.47 & 0.54 & 0.41 \\
\addlinespace
NMN w/  Graph-count + decont.                   & 67.3 & 0.47 & 0.70 & 0.45 & 0.42 & 0.47 & 0.55 & 0.33 \\

\addlinespace
NMN w/ Graph-count + pretraining            & 69.6 & 0.58 & 0.70 & 0.47 & 0.42 & 0.49 & 0.58 & 0.41 \\
\addlinespace
NMN w/ Graph-count + decont. + pretraining        & 68.7 & 0.58 & \textbf{0.79} & \textbf{0.55} & \textbf{0.54} & \textbf{0.55} & 0.62 & 0.43 \\
\bottomrule
\end{tabular}
}
\caption{Faithfulness scores on \nlvr{} using the average over module occurrences evaluation.}
\label{tab:nlvr-results-method-2}
\end{table*}
\begin{table*}[t]
\small
\centering
\captionsetup{font=footnotesize}
\resizebox{1.0\textwidth}{!}{
\begin{tabular}{lcccccccc}
\toprule
\multirow{2}[3]{*}{{\bf Model}} & \multirow{2}[3]{*}{\begin{tabular}{@{}c@{}}\textbf{Performance} \\ (Accuracy) \end{tabular}} & \multicolumn{3}{c}{\textbf{Overall Faithful.}($\uparrow$)}    & \multicolumn{4}{c}{Module-wise Faithfulness($\uparrow$)} \\
\cmidrule(lr){3-9}
    &  & Prec. & Rec. & F1 & find & filter & with-relation & relocate \\
\midrule
LXMERT                               & \textbf{71.7} &      &      &      &      &      \\
\addlinespace
Upper Bound                   &      & 1 & 0.8377 & 0.89    &  0.89    &  0.92    &    0.95  &   0.75   \\
\addlinespace
NMN w/ Layer-count                        & 71.2 & 0.59 & 0.39 & 0.25 & 0.31 & 0.28 & 0.45 & 0.30 \\
NMN w/ Sum-count                          & 68.4 & \textbf{0.79} & 0.31 & 0.34 & 0.38 & 0.36 & 0.48 & 0.28 \\
NMN w/ Graph-count                        & 69.6 & 0.68 & 0.39 & 0.38 & 0.43 & 0.36 & 0.44 & 0.22 \\
\addlinespace
NMN w/  Graph-count + decont.                   & 67.3 & 0.62 & 0.51 & 0.47 & 0.53 & 0.39 & 0.43 & 0.16 \\

\addlinespace

NMN w/ Graph-count + pretraining            & 69.6 & 0.70 & 0.49 & 0.47 & 0.52 & 0.41 & 0.51 & 0.27 \\
\addlinespace

NMN w/ Graph-count + decont. + pretraining        & 68.7 & 0.71 & \textbf{0.66} & \textbf{0.62} & \textbf{0.68} & \textbf{0.50} & \textbf{0.55} & \textbf{0.31} \\
\bottomrule
\end{tabular}
}
\caption{Faithfulness scores on \nlvr{} using a negative IOU threshold of $10^{-8}$ and example-wise averaging.}

\label{tab:nlvr-results-iou-minus-1e-8}
\end{table*}

\section{Details about Experiments}
\label{app:tr-details}
\paragraph{Visual Reasoning} We use the published pre-trained weights and the same training configuration of LXMERT \cite{lxmert}, with 36 bounding boxes proposed per image. Due to memory constraints, we restrict training data to examples having a gold program with at most 13 modules.

\subsection{Program Annotations}
\label{app:annotation}
We generated program annotations for NLVR2 by automatically canonicalizing its question decompositions in the \textsc{Break} dataset \cite{Wolfson2020Break}. Decompositions were originally annotated by Amazon Mechanical Turk workers. For each utterance, the workers were asked to produce the correct decomposition and an \emph{utterance attention} for each operator (module), whenever relevant.

\begin{figure}[tb]
    \centering
    \includegraphics[width=70mm]{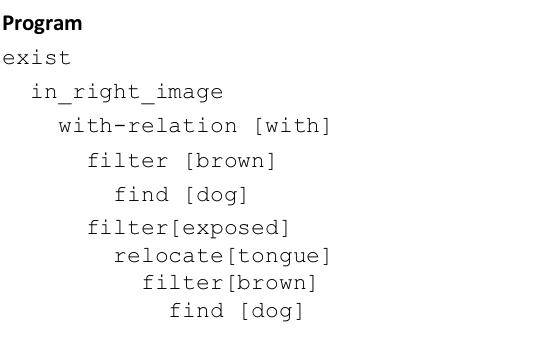}
    \caption{An example of a gold program for NLVR2 that is unnecessarily complicated.}
    \label{fig:bad_program}
\end{figure}

\paragraph{Limitations of Program Annotations} Though our annotations for gold programs in NLVR2 are largely correct, we find that there are some examples for which the programs are unnecessarily complicated. For instance, for the sentence ``the right image contains a brown dog with its tongue extended.'' the gold program is shown in Figure \ref{fig:bad_program}. This program could be simplified by replacing the \texttt{with-relation} with the second argument of \texttt{with-relation}. Programs like this make learning more difficult for the NMNs since they use modules (in this case, \texttt{with-relation}) in degenerate ways. There are also several sentences that are beyond the scope of our language, e.g. comparisons such as ``the right image shows exactly two virtually identical trifle desserts.''

\section{Significance tests}
\subsection{Visual Reasoning}
\label{app:sig-test-visual}
We perform a paired permutation test to test the hypothesis $ H_0 $: \emph{NMN w/ Graph-count + decont. + pretraining} has the same inherent faithfulness as \emph{NMN w/ Layer-count}. We follow the procedure described by \citet{ventura-sig-test}, which is similar to tests described by \citet{yeh-2000-more} and \citet{noreen1989computer}. Specifically, we perform $ N_{total} = 100,000 $ trials in which we do the following. For every example, with probability $1/2$ we swap the F$_1$ scores obtained by the two models for that example. Then we check whether the difference in the aggregated F$_1$ scores for the two models is at least as extreme as the original difference in the aggregated F$_1$ scores of the two models. The p-value is given by $ N_{exceed}/N_{total} $, where $ N_{exceed} $ is the number of trials in which the new difference is at least as extreme as the original difference.

\end{document}